\documentclass[letterpaper, 10 pt, conference]{ieeeconf}  

\IEEEoverridecommandlockouts                              

\usepackage{graphics} 
\usepackage{graphicx}
\usepackage{subcaption}
\usepackage{mathptmx} 
\usepackage{amsmath} 
\usepackage{amssymb}  
\usepackage{dsfont}
\usepackage{algorithm}
\usepackage{algorithmicx}
\usepackage[noend]{algpseudocode}

\RequirePackage{xspace}
\makeatletter
\DeclareRobustCommand\onedot{\futurelet\@let@token\@onedot}
\def\@onedot{\ifx\@let@token.\else.\null\fi\xspace}

\def\ie{\emph{i.e}\onedot}

\def\etc{\emph{etc}\onedot}

\def\etal{\emph{et al}\onedot}
\makeatother

\title{\LARGE \bf
Multiple Rotation Averaging with Constrained Reweighting Deep Matrix Factorization
}

\author{Shiqi Li$^{1}$, Jihua Zhu$^{1}$, Yifan Xie$^{1}$, Naiwen Hu$^{1}$, Mingchen Zhu$^{2}$, Zhongyu Li$^{1}$ and Di Wang$^{1}$
\thanks{$^{1}$School of Software Engineering, Xi’an Jiaotong University.}
\thanks{$^{2}$University of California, Davis.}
\thanks{Jihua Zhu is the correspondence author.}
}

\begin{document}

\maketitle
\thispagestyle{empty}
\pagestyle{empty}

\begin{abstract}

Multiple rotation averaging plays a crucial role in computer vision and robotics domains. The conventional optimization-based methods optimize a nonlinear cost function based on certain noise assumptions, while most previous learning-based methods require ground truth labels in the supervised training process. Recognizing the handcrafted noise assumption may not be reasonable in all real-world scenarios, this paper proposes an effective rotation averaging method for mining data patterns in a learning manner while avoiding the requirement of labels. Specifically, we apply deep matrix factorization to directly solve the multiple rotation averaging problem in unconstrained linear space. For deep matrix factorization, we design a neural network model, which is explicitly low-rank and symmetric to better suit the background of multiple rotation averaging. Meanwhile, we utilize a spanning tree-based edge filtering to suppress the influence of rotation outliers. What's more, we also adopt a reweighting scheme and dynamic depth selection strategy to further improve the robustness. Our method synthesizes the merit of both optimization-based and learning-based methods. Experimental results on various datasets validate the effectiveness of our proposed method.

\end{abstract}

\section{INTRODUCTION}
\label{sec:introduction}

Multiple rotation averaging (MRA)~\cite{arrigoni2020synchronization}, which is also known as rotation synchronization, is a fundamental problem in 3D computer vision and robotics. It finds wide application in various domains such as Structure from Motion (SfM)~\cite{cui2015global,chen2021hybrid}, pose graph optimization (PGO) in visual Simultaneous Localization and Mapping (SLAM) system~\cite{mur2015orb}, multiview point cloud registration~\cite{gojcic2020learning,wang2023robust,li2024matching}, and sensor network localization~\cite{tron2009distributed}.  

The MRA problem belongs to the broader category of group synchronization problems, specifically centered on the Special Orthogonal Group $SO(3)$. The primary goal of this task is to estimate the absolute orientation of individual nodes (camera, Lidar, \etc) from the relative rotation measurements between them. Typically, the relative measurements in practice are contaminated by the noise and outlier introduced by the imperfect pairwise matching or registration, which further makes the problem more challenging. 

The majority of previous methods formulate the MRA as an optimization problem~\cite{hartley2011l1,chatterjee2017robust,shi2020message}. In order to satisfy the condition of nonlinear $SO(3)$ space, these methods typically need to impose extra constraints during the optimization. This challenging process often entails local linearization or iterative optimization techniques. Furthermore, addressing outliers requires the application of robust loss functions that rely on assumptions about sensor noise and outlier distribution. However, these assumptions may not accurately reflect real-world conditions. Moreover, it frequently involves manually adjusting parameters, a task that can be nontrivial.  

Recent learning-based methods~\cite{purkait2020neurora,yew2021learning,li2022rago} have harnessed the power of neural networks to address the MRA problem. These methods employ Graph Neural Network (GNN) to model the pose graph. Relative rotations and absolute orientations are encoded as edge messages and node features. Diverse message passing and aggregation schemes are designed for learning noise and outlier distributions from training data. While these supervised methods demonstrate impressive performance, they continue to grapple with data scarcity and domain gaps.

In this study, we directly recover the absolute orientation of every node in linear space via the deep matrix factorization technique. All target absolute orientations are stacked and represented as a product of several factor matrices using a linear neural network. The network is deliberately designed to possess an explicitly low-rank and symmetric structure, aligning with the background of the MRA problem. We employ the observed relative measurements as constraints for network optimization to facilitate unsupervised learning. Additionally, we employ a spanning tree-based edge filtering and reweighting scheme to enhance the robustness. In contrast to traditional optimization-based methods, our approach embraces a data-driven method to extract information from observed relative measurements, significantly reducing the need for noise distribution assumption and manual adjustment of threshold parameters. Despite being a learning-based approach, our method is distinct from previous works in that it is unsupervised, liberating us from the laborious dataset acquisition. Furthermore, our method eliminates the need to address adaptation challenges across different domains. Finally, unlike previous work~\cite{tejus2023rotation}, we apply more sophisticated designs based on prior knowledge to considerably aid the learning process and bypass eigen decomposition to retrieve the absolute orientations directly. Extensive experiments on datasets with various scales and modalities demonstrate the accuracy and generalization ability of our proposed method and our code will be available upon acceptance.

In conclusion, our contribution can be summarized as: 1) We introduce a spanning tree-based edge filtering to effectively and efficiently detect and remove outliers in relative measurements. 2) We design a network with explicit low-rank and symmetric constraints to directly calculate absolute orientations in unrestricted linear space using deep matrix factorization. 3) We propose a reweighing scheme and a depth selection mechanism to enhance the robustness.

\section{Related work}
\label{sec:related_work}

\paragraph{Conventional MRA Method}
Multiple rotation averaging is an extensively researched problem that was first introduced by Govindu and addressed using a linear model in~\cite{govindu2001combining}. To obtain more accurate results, many approaches iteratively optimize robust cost functions to suppress the impact of outliers. Hartley \etal~\cite{hartley2011l1} employ single rotation averaging under the $\ell_1$ norm to update absolute orientations and introduce a rotation averaging method based on the Weiszfeld algorithm. Chatterjee and Govindu~\cite{chatterjee2017robust} present a method that initialized with a spanning tree and then uses an iteratively reweighted least squares (IRLS) procedure to minimize $\ell_{\frac{1}{2}}$ loss, which yields the best empirical accuracy. Arrigoni \etal~\cite{arrigoni2018robust} likewise employ the IRLS technique to bolster the robustness of the spectral decomposition method outlined in~\cite{arie2012global}. Shi and Lerman~\cite{shi2020message} introduce a message passing least squares (MPLS) framework as an alternative approach to IRLS. Zhang \etal~\cite{zhang2023revisiting} propagate the uncertainty from the correspondences into the rotation averaging to better model the underlying noise distributions. 

Besides, some approaches concentrate on spotting outliers within the input graph. Govindu~\cite{govindu2006robustness} introduces a RANSAC-style sampling approach to eliminate outliers. Zach \etal~\cite{zach2010disambiguating} present a Bayesian framework for classifying edges as either inliers or outliers. Arrigoni \etal~\cite{arrigoni2014robust} propose heuristic methods based on cycle bases to improve cycle consistency check. HARA~\cite{lee2022hara} introduces a heuristic method that incrementally constructs a spanning tree based on triplet support. 

Additionally, some incremental estimation-based approaches also demonstrate impressive performance. Among them, the IRA~\cite{gao2021incremental} series is particularly representative.

\paragraph{Learning-based MRA Method}
In recent years, deep learning-based algorithms have been proposed to tackle the multiple rotation averaging problem. NeuRoRA~\cite{purkait2020neurora} pioneers neural network architectures on multiple rotation averaging, the proposed network is a combination of cleaning and fine-tuning sub-networks. LITS~\cite{yew2021learning} proposes a weight-shared message passing neural network to predict an incremental in each iteration. PoGO-Net~\cite{li2021pogo} designs a de-noising layer to achieve implicit edge-dropping. MSP~\cite{yang2021end} integrates differentiable multiple sources initialization and optimization procedure in a unified GNN. RAGO~\cite{li2022rago} decouples the MRA to multiple single rotation averaging problems and use a gated recurrent unit module to enhance robustness. The majority of these solutions are based on the GNN paradigm and need ground truth orientation labels to supervise network training. However, collecting accurate realistic MRA dataset is tricky, and such data scarcity will impede the development of these supervised methods. 

A recent work, DMF-SYNCH~\cite{tejus2023rotation}, proposes to utilize the deep matrix factorization to solve the multiple rotation averaging problem in an unsupervised way. While there is still a lot of room for improvement in performance, this method provides a new view for multiple rotation averaging. In this paper, we use a similar technique but integrate additional prior knowledge to solve the problem more straightforwardly. 

\section{Preliminary}
\label{sec:preliminary}

\subsection{Notation and Problem Statement}
\label{subsec:notation}
A rotation is denoted by a rotation matrix $\mathbf{R}$, the Riemannian distance between two rotations $\mathbf{R}_1$ and $\mathbf{R}_2$ can be straightforwardly determined as the angular difference between them, \ie,
\begin{equation}
    d(\mathbf{R}_1, \mathbf{R}_2)=\Vert\text{Log}(\mathbf{R}_1\mathbf{R}_2^\top)\Vert,
\end{equation}
where $\text{Log}(\cdot)$ is logarithm map function and $\Vert\cdot\Vert$ is the Euclidean norm. 

We also introduce the chordal distance, which is akin to the Riemannian distance when two rotations are proximate to each other. The chordal distance between $\mathbf{R}_1$ and $\mathbf{R}_2$ is defined as:
\begin{equation}
    d_{chord}(\mathbf{R}_1, \mathbf{R}_2) = \Vert\mathbf{R}_1 - \mathbf{R}_2\Vert_F,
\end{equation}
where $\Vert\cdot\Vert_F$ represents the Frobenius norm. 

Given a 3D scene with multiple frames, each frame can be considered as a camera in SfM or a point cloud in multiview registration, the geometric relationships within the 3D scene can be represented by a view-graph $\mathcal{G}=\{\mathcal{V}, \mathcal{E}\}$. Each vertex $v_i$ in $\mathcal{V}$ stands for a frame with unknown absolute orientation $\mathbf{R}_i$, the cardinality of vertex set $|\mathcal{V}|$, i.e., the number of frames is denoted as $N$. An edge $e_{ij}$ in $\mathcal{E}$ indicates the existence of pairwise relative rotation $\mathbf{R}_{ij}$ between vertex $v_i$ and $v_j$. Ideally, the relative rotation and absolute orientation should satisfy the constraint that $\mathbf{R}_{ij}=\mathbf{R}_i\mathbf{R}_j^\top$. However, this condition is often not met due to the noise and outliers in practical relative measurement $\mathbf{\tilde{R}}_{ij}$. 

Consequently, we derive the definition of multiple rotation averaging as seeking a set of absolute orientation $\mathbf{R}_\mathcal{V}$ that best align with the observed relative rotations, \ie,
\begin{equation}
    \mathbf{R}_\mathcal{V} = \underset {\{\mathbf{R}_1,\mathbf{R}_2,\cdots,\mathbf{R}_N\}} { \operatorname {arg\,min} }  \sum_{e_{ij}\in\mathcal{E}} d(\mathbf{\tilde{R}}_{ij}-\mathbf{R}_i\mathbf{R}_j^\top)^p, 
    \label{eq:problem}
\end{equation}
where $p$ is typically set to 1 or 2. 

\subsection{Eigen Decomposition and Matrix Reconstruction}
\label{subsec:decomposition}
Let's start with a simpler case. Assume $\mathcal{G}$ is a complete graph, with all relative rotations known. We can construct a $3N\times 3N$ block matrix $\mathbf{G}$ by concatenating all pairwise rotation matrices. Additionally, we create a $3N\times 3$ matrix $\mathbf{X}$ by stacking all absolute orientations,
\begin{equation}
    \mathbf{G}=
    \begin{bmatrix}
        \mathbf{I} & \mathbf{R}_{12} & \cdots & \mathbf{R}_{1N} \\
        \mathbf{R}_{21} & \mathbf{I} & \cdots & \mathbf{R}_{2N} \\
        \vdots & \vdots & \ddots & \vdots \\
        \mathbf{R}_{N1} & \mathbf{R}_{N2} & \cdots & \mathbf{I}
    \end{bmatrix},
    \mathbf{X}=
    \begin{bmatrix}
        \mathbf{R}_1 \\
        \mathbf{R}_2 \\
        \vdots \\
        \mathbf{R}_N
    \end{bmatrix}.
\end{equation}

Given that $\mathbf{G}$ can be expressed by $\mathbf{X}$ (\ie, $\mathbf{G}=\mathbf{X}\mathbf{X}^\top$), we can deduce that matrix $\mathbf{G}$ has a rank of 3. To estimate $\mathbf{X}$ from $\mathbf{G}$, we can operate eigen decomposition on $\mathbf{G}$ and stack three eigenvectors to get a $3N\times 3$ matrix, then we project each $3\times3$ submatrix to $SO(3)$ to obtain valid orientation. For more details, we refer to the original work~\cite{arie2012global}. 

Based on the eigen decomposition method, we can cast the MRA problem as a matrix reconstruction problem~\cite{arrigoni2014robustb}, which involves recovering the matrix $\mathbf{G}$ from an incomplete matrix $\mathbf{\tilde{G}}$. 
In the context of practical MRA problem, the observed matrix $\mathbf{\tilde{G}}$ often contains noise and outliers. This necessitates the method not only completing the missing parts but also recognizing and fixing these disturbances, rendering the problem more challenging.

\section{Method}
\label{sec:method}
\begin{figure*}[t]
  \centering
    \includegraphics[width=\linewidth]{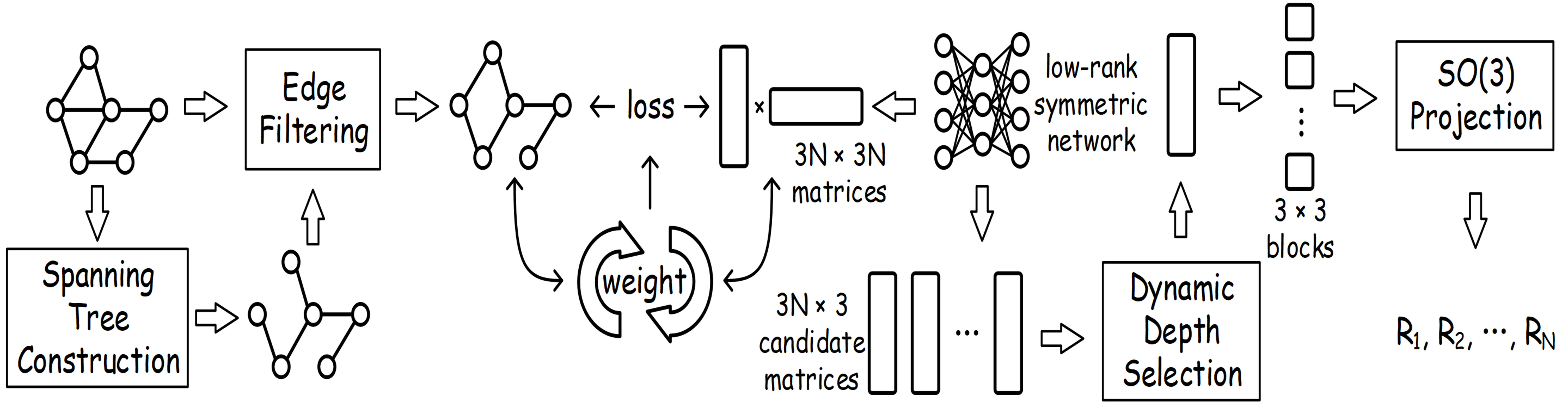}
    \caption{Pipeline of our proposed method. It starts from the construction of spanning tree, which is utilized to remove pairwise rotation outliers. Under the reweighting scheme, the constrained neural network is optimized by reliable pairwise rotations. Finally, network parameters that best aligns with the observations are selected to perform the projection operation and generate outputs.}
   \label{fig:pipeline}
\end{figure*}

Fig. \ref{fig:pipeline} illustrates the pipeline of our method, which includes the spanning tree-based edge filtering and a constrained neural network with reweighting scheme. The spanning tree is constructed from all pairwise rotations to filter rotation outliers. Under the reweighting scheme, reliable pairwise rotations are used as supervision of a constrained neural network to compute absolute orientations with defined loss function.

\subsection{Spanning Tree-based Edge Filtering}
\label{subsec:filtering}

\renewcommand{\algorithmicrequire}{\textbf{Input:}}  
\renewcommand{\algorithmicensure}{\textbf{Output:}} 
\begin{algorithm}[ht]
    \caption{Spanning tree-based edge filtering}
    \begin{algorithmic}[1]
    \Require pose graph $\mathcal{G}=\{\mathcal{V}, \mathcal{E}\}$ and filter threshold $\sigma$
    \Ensure filtered pose graph $\mathcal{G'}$
    \State set $\Phi=\{\}$
    \For{$e_{ij}$ in $\mathcal{E}$} \Comment{Enumerate triplet loops}
        \State set $\Theta_{ij}=\{\}$
        \For{$k$ in $\mathcal{V}$}
            \If{$e_{ik}\in\mathcal{E}$ and $e_{kj}\in\mathcal{E}$}
                \State add $\delta_{ij,k}$ to $\Theta_{ij}$ and $\Phi$
            \EndIf
        \EndFor
    \EndFor
    \State $\epsilon=\text{median}(\Phi)$
    \For{$e_{ij}$ in $\mathcal{E}$} \Comment{Assign edge attributes}
        \State $\delta_{ij}=\sum\Theta_{ij}/|\Theta_{ij}|$ and $s_{ij}=0$
        \For{$\delta_{ij,k}$ in $\Theta_{ij}$}
            \State $s_{ij}=s_{ij}+\mathds{1}(\delta_{ij,k}<\epsilon)$
        \EndFor
    \EndFor
    \State run greedy algorithm to build tree and generate $\mathbf{R}^{tree}$
    \For{$e_{ij}$ in $\mathcal{E}$} \Comment{Delete non-conforming edges}
        \If{$d_{chord}(\mathbf{\tilde{R}}_{ij}, \mathbf{R}_{ij}^{tree})>\sigma$}
            \State remove $e_{ij}$ from $\mathcal{E}$
        \EndIf
    \EndFor
    \end{algorithmic}
    \label{alg:filtering}
\end{algorithm}

The pairwise rotations are usually provided by the upstream procedure, which are inevitably noisy. Although the further deep matrix factorization module is capable of denoising the input pose graph, it can still benefit from detection and exclusion of incorrect relative rotations.

To identify the outliers, we first construct a spanning tree of the input pose graph. This spanning tree provides pairwise rotation estimates between arbitrary vertices, and then we eliminate observed rotations that are inconsistent with the tree. Constructing the tree necessitates the assignment of attributes to each edge and the definition of comparison operations between them. With attributes and comparative relationships, we can apply the minimum spanning tree algorithm to construct the desired tree. The design of the edge attributes directly determines the selection of the final tree. We aim to assign attributes that entitle inlier edges with high priority and outlier edges with lower priority. In pose graph data, loop consistency serves as the most intuitive way for identifying anomalous edges. In this work, for simplicity, we utilize a triple loop. Consider three vertices $i$, $j$, and $k$, where all relative rotations between them are known, we define the loop error composed of edge $e_{ij}$ and vertex $k$ as:
\begin{equation}
    \delta_{ij,k}=d_{chord}(\mathbf{R}_{ij}, \mathbf{R}_{ik}\mathbf{R}_{kj}).   
\end{equation}

In an ideal, noise-free setting, for any loop, the $\delta$ should be close to zero. However, when at least one edge in the loop is an outlier, such deviation has a high probability of being significant. We use statistical information about the error $\delta$ among the graph to establish the attributes of each edge. Specifically, for each edge $e_{ij}$, we collect all loop errors composed by it to $\Theta_{ij}$, and the mean of $\Theta_{ij}$ is assigned to edge $e_{ij}$ as an attribute. However, the topological structure of input pose graph is unknown, and the degree of each vertex is typically not normally distributed. Therefore, using only the mean of loop errors is inappropriate, so we further count the support count $s_{ij}$ of each edge, \ie,
\begin{equation}
    s_{ij}=\sum_{\delta_{ij,k}\in\Theta_{ij}}\mathds{1}(\delta_{ij,k}<\epsilon),
\end{equation}
where $\mathds{1}(\cdot)$ is indicator function. The threshold $\epsilon$ is set to the median of all triplet loop errors in the graph, making the method adaptable to varying input data. 

During the edge comparison, we first prioritize edges based on their support counts, with higher $s$ values receiving higher priority. In cases where edges have the same support count, we arrange them in ascending order of mean error. Based on the comparison operations defined above, we can employ the Prim~\cite{prim1957shortest} or Kruskal~\cite{kruskal1956shortest} algorithm to construct the spanning tree. 

Once we have obtained the desired spanning tree, it can be used to filter the outliers in the input pose graph. Following ~\cite{lee2022hara}, we eliminate edges that do not conform with the relative rotations $\mathbf{R}^{tree}$ provided by the tree, \ie,
\begin{equation}
    d_{chord}(\mathbf{\tilde{R}}_{ij}, \mathbf{R}_{ij}^{tree})>\sigma,
\end{equation}
where $\sigma$ is a predefined threshold.  

Alg. \ref{alg:filtering} summarizes our edge filtering method.

\subsection{Deep Matrix Factorization}
\label{subsec:factorization}
As discussed in Sec. \ref{sec:preliminary}, we have transformed the multiple rotation averaging into a matrix reconstruction problem. Here, we elucidate vanilla matrix reconstruction via deep matrix factorization and subsequently derive our proposed solution, which is more straightforward and applicable to the MRA problem.

The objective of the matrix reconstruction is to recover the complete matrix from a partially observed subset. Directly solving this problem is infeasible due to the existence of infinitely many valid solutions. To render this under determined problem manageable, certain assumptions must be introduced. Typically, matrix reconstruction is carried out under the assumption of low-rank. The low-rank assumption has a number of beneficial characteristics. For instance, it encourages cooperative relationships between columns and rows, which aids in understanding. In the context of our multiple rotation averaging problem, the low-rank assumption is especially suitable. This is due to our prior knowledge that the ground truth matrix $\mathbf{G}$ has a rank of 3. Thus, our matrix reconstruction problem can be formulated as the task of finding a low-rank matrix $\mathbf{\hat{G}}$ based on the observation $\mathbf{\tilde{G}}$, 
\begin{equation}
    \begin{split}
        \text{minimize} &\quad \text{rank}(\mathbf{\hat{G}}) \\
        \text{subject to} &\quad \mathbf{\hat{G}}\odot\Omega\approx\mathbf{\tilde{G}}\odot\Omega,
    \end{split}
    \label{eq:objective}
\end{equation}
where $\odot$ represents the Hadamard product, $\Omega$ is the binary mask which has the same dimensions as $\mathbf{G}$, when $\mathbf{G}_{ij}$ is available, the $\Omega_{ij}$ is a $3 \times 3$ block of ones, otherwise, $\Omega_{ij}$ is a $3 \times 3$ block of zeros.   

Directly optimizing objective (\ref{eq:objective}) is challenging due to the presence of the rank term, which renders it a non-convex optimization problem. Most recent studies transform the original problem into convex optimization using convex relaxation techniques~\cite{hu2012fast}, such as optimizing the nuclear norm of $\mathbf{\hat{G}}$. Instead of adopting an explicit low-rank approximation, some other studies explore implicit regularization techniques to obtain a low-rank solution~\cite{gunasekar2017implicit,arora2019implicit}. Here, we adhere to this paradigm by multiplying a number of factors $\mathbf{W}$ to recreate the entire matrix. This technique is also referred to as deep matrix factorization. Formally, the reconstructed matrix $\mathbf{\hat{G}}$ can be expressed as: 
\begin{equation}
    \mathbf{\hat{G}}=\prod_{i=1}^d\mathbf{W}_i,
\end{equation}
where $d$ is referred to as the depth of the factorization.

Previous observations indicate that when using small enough step sizes and initialization close to the origin, gradient descent-based methods tend to converge toward low-rank solutions~\cite{gunasekar2017implicit}. This implicit bias, which can be seen as a form of regularization, introduces a novel approach to the matrix reconstruction problem. Therefore, we can optimize the following problem to estimate $\mathbf{G}$,  
\begin{equation}
    \text{minimize} \quad \rho(\mathbf{\hat{G}}\odot\Omega - \mathbf{\tilde{G}}\odot\Omega),
    \label{eq:minimize}
\end{equation}
where $\rho(\cdot)$ is some cost function.

Problem (\ref{eq:minimize}) can be readily implemented in prevalent deep learning frameworks by a sequence of linear layers (without bias term) omitting activation function. Forward propagation naturally yields the complete matrix estimation, and the factor matrices (the weights of linear layers) can be conveniently updated during backward propagation.

\begin{figure}[t]
  \centering
    \begin{subfigure}{0.45\linewidth}
        \centering
        \includegraphics[width=\linewidth]{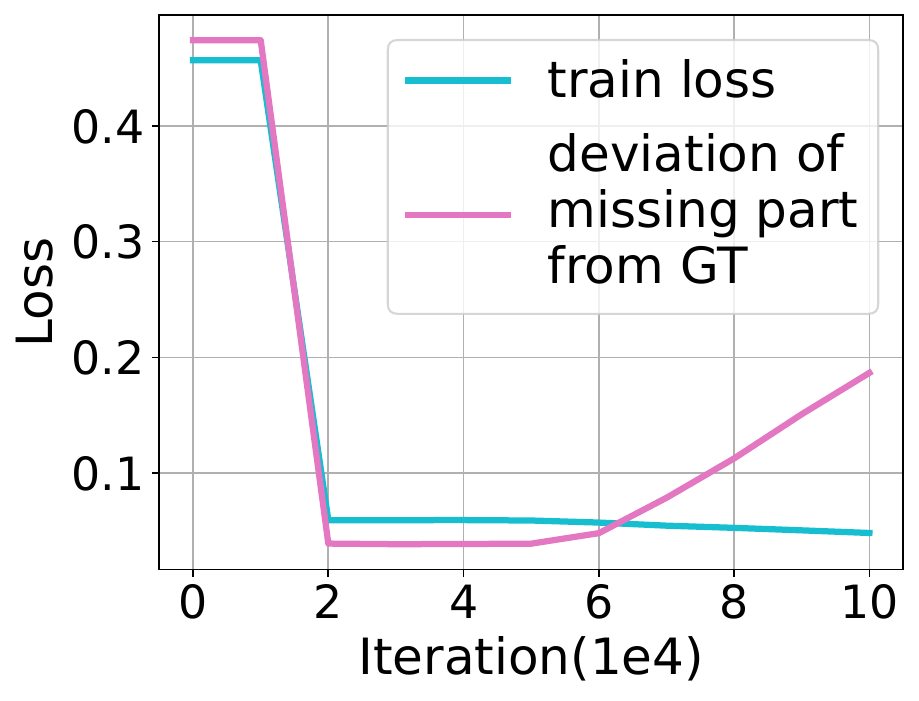}
        \caption{Loss.}
    \end{subfigure}
    \begin{subfigure}{0.52\linewidth}
        \centering
        \includegraphics[width=\linewidth]{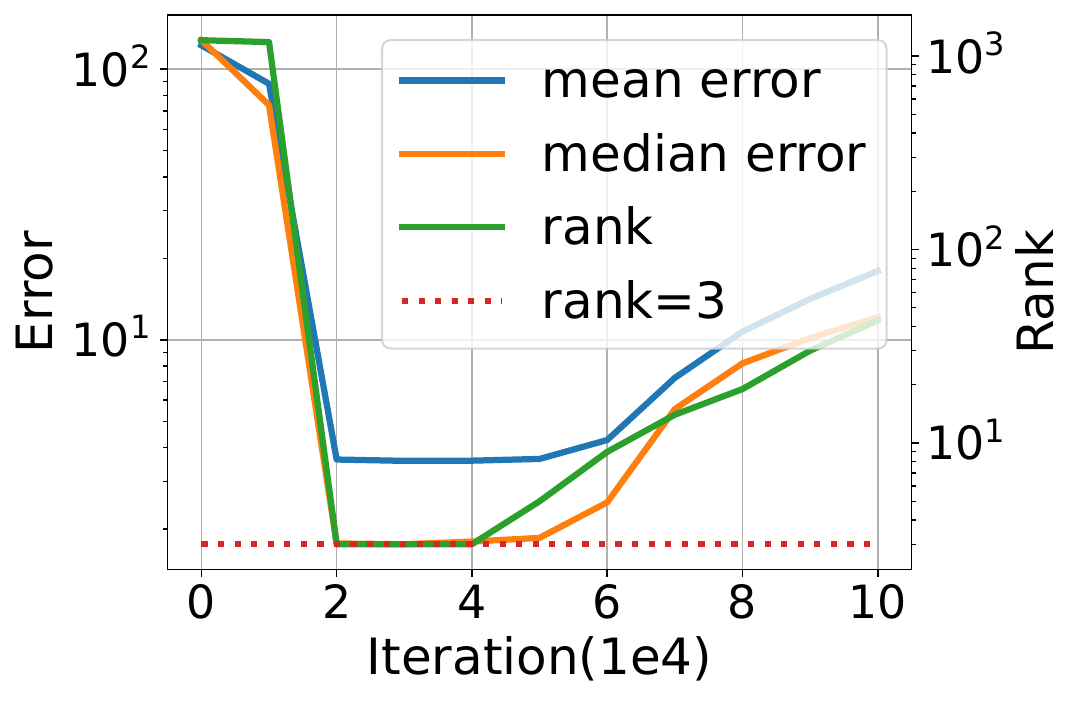}
        \caption{Error and rank.}
    \end{subfigure}
    \caption{Training process of a linear network model without explicit low-rank constraint.}
   \label{fig:iteration}
\end{figure}

Although implicit regularization tends to favor low-rank solutions, we argue that it would be better to explicitly constrain the rank of the product matrix. On one hand, unlike fields such as image processing or recommendation systems, which rely on a low-rank assumption but lack access to the actual rank of the target full matrix, in our multiple rotation averaging problem, as discussed in Sec. \ref{subsec:decomposition}, we have knowledge that the rank of the ground truth matrix $\mathbf{G}$ is 3. On the other hand, we have observed that the implicit regularization from gradient-based optimization is insufficient to impose a strong constraint on low-rank solutions, particularly in noisy scenarios. In~\cite{arora2019implicit,tejus2023rotation}, a vanilla model, in which the dimensions of the factors coincide with the dimensions of $\mathbf{G}$ (\ie, $\forall\mathbf{W}\in\mathbb{R}^{3N\times3N}$) is employed. The training procedure of a vanilla network with 5 linear layers on the Yorkminster scene from the 1DSfM dataset~\cite{wilson2014robust} is depicted in Fig. \ref{fig:iteration}. We can observe that as the training iterations increase, the training loss consistently decreases. However, the error and rank of the estimated matrix initially decrease but then rapidly increase. This overfitting phenomenon limits the practical applicability of the deep matrix factorization-based method for the multiple rotation averaging problem in real-world applications.

Since \emph{(1)} we already know that the true rank value of $\mathbf{G}$ is 3, and \emph{(2)} existing implicit regularization methods suffer from overfitting, we implement a bottleneck design in our linear neural network to impose an explicit constraint. Instead of considering some nuclear norms, we achieve explicit low-rank by directly restricting the shared dimension of factor matrices. Specifically, we set   
\begin{equation}
    \mathbf{W}_{\frac{d}{2}}\in\mathbb{R}^{3N\times3} \quad \text{and} \quad \mathbf{W}_{\frac{d}{2}+1}\in\mathbb{R}^{3\times3N},
\end{equation}
in our linear layer sequence, based on the propriety that, 
\begin{equation}
    \text{rank}(\mathbf{\hat{G}})=\text{rank}(\prod_{i=1}^d\mathbf{W}_i)\leq \min\text{rank}(\mathbf{W}_i).
\end{equation}

This straightforward yet effective design imposes a stronger constraint that prevents the network from overfitting the noise and outliers in pairwise rotations. In addition to the certain rank, the ground truth $\mathbf{G}$ possesses another property: it is symmetric and positive semidefinite. As a result, we only need to optimize half of the parameters, since the former and latter halves can share them,
\begin{equation}
    \mathbf{\hat{G}}=\mathbf{H}\mathbf{H}^\top \quad \text{where} \quad \mathbf{H}=\prod_{i=1}^{d/2}\mathbf{W}_i.
    \label{eq:symmetric}
\end{equation}

The Eq. \ref{eq:symmetric} is actually analogous to the definition of the matrix $\mathbf{G}$, and the product term $\mathbf{H}$ can be considered an estimate of stacked orientations (\ie, matrix $\mathbf{X}$ in Sec. \ref{subsec:decomposition}). Such clear semantics further enhance the interpretability of our model. Furthermore, thanks to the symmetric design, reconstructing matrix $\mathbf{G}$ and performing eigen decomposition are no longer demanded.  We can simply project the blocks in $\mathbf{H}$ into the $SO(3)$ space to obtain the absolute orientations.

Despite the similarity in formulation, our method differs fundamentally from DMF-SYNCH~\cite{tejus2023rotation}. Deep matrix decomposition is utilized in~\cite{tejus2023rotation} as an upgrade of Alternating Direction Method of Multipliers (ADMM) to achieve superior reconstruction of $\mathbf{G}$, while the entire pipeline remains consistent with the paradigm in~\cite{arrigoni2014robustb}. However, our technique directly expresses the orientation matrix as a product of factor matrices, demonstrating that MRA can be optimized in linear space while disregarding the $SO(3)$ manifold restrictions. 

\begin{figure}[t]
    \centering
    \begin{subfigure}{0.49\linewidth}
        \centering
        \includegraphics[width=\linewidth]{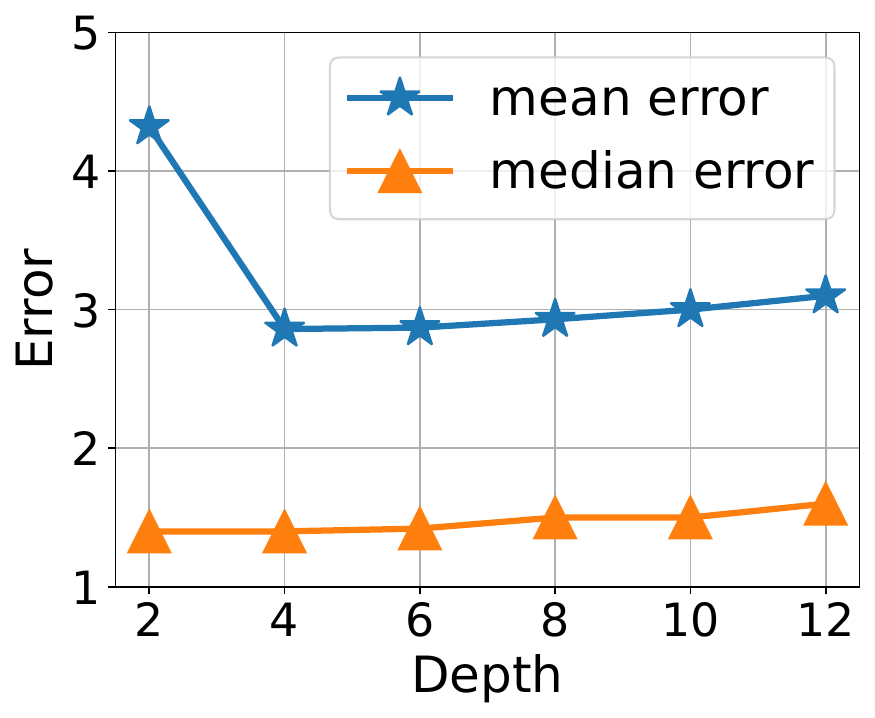}
        \caption{Yorkminster.}
    \end{subfigure}
    \begin{subfigure}{0.49\linewidth}
        \centering
        \includegraphics[width=\linewidth]{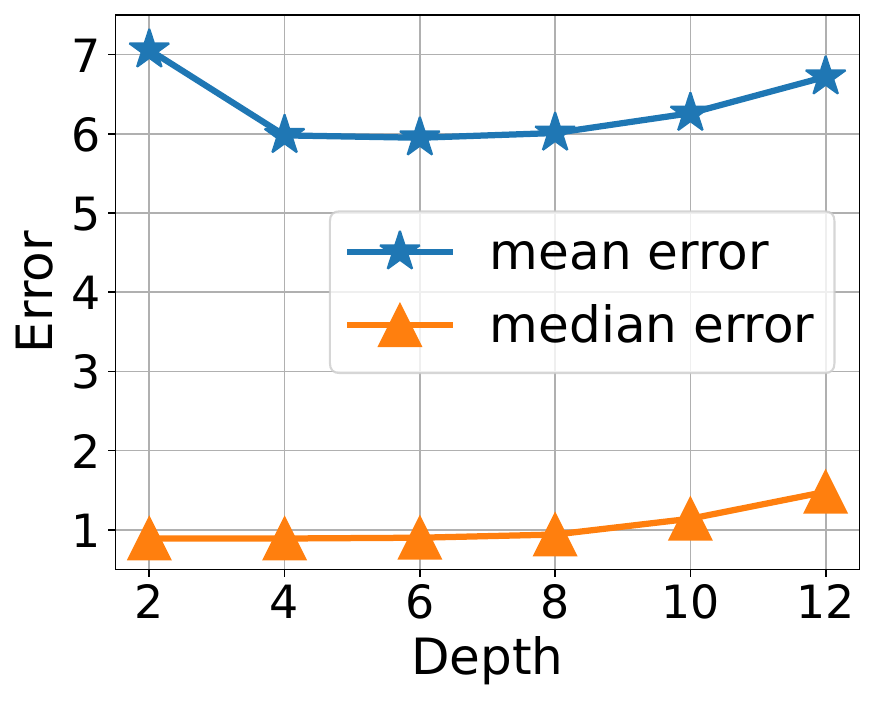}
        \caption{Madrid Metropolis.}
    \end{subfigure}
    \caption{Errors of models with different depth settings.}
   \label{fig:depth}
\end{figure}

Another noteworthy parameter is the depth of the factorization. Previous studies have suggested that increasing the depth leads to more accurate results~\cite{arora2019implicit,tejus2023rotation}. However, in our experiments, we have observed that adding excessive depth can negatively impact performance. Fig. \ref{fig:depth} illustrates the results on two scenarios from 1DSfM dataset using different depth settings. It is evident that increasing the depth from 2 to 4 results in a noticeable improvement. However, when the depth exceeds 8, there is a decline in performance. In our experiments, we discovered that in most scenarios, a network with 4 to 8 layers yields satisfactory results. To further determine the appropriate depth, we introduce a flexible hypothesis evaluation mechanism for dynamic depth selection. Formally, we build multiple candidate models with depths ranging from 2 to 8 and optimize them in the same way. These candidate models are independent of one another and can be individually optimized in a distributed fashion. Finally, the model that best fits the observation (using Eq. \ref{eq:problem} as a discriminator) will be selected to provide final results.

\subsection{Reweighting Scheme}
\label{subsec:reweighing}
To improve the resilience of our model, we provide an iterative reweighting scheme inspired by IRLS. As an important technique in the field of robust estimation, IRLS identifies outliers within the data distribution and assigns fresh weights in each iteration to mitigate the their influence~\cite{huang2017translation,huang2019learning}. Extending this idea, we assign weights to each observed rotation matrix and adjust them continuously while optimizing. By decreasing weights to noisy rotations, our scheme effectively makes their influence negligible.

Formally, in reweighting step $t$, we calculate the chordal distance between estimated matrix $\mathbf{\hat{G}}$ and the observed $\mathbf{\tilde{G}}$ in available parts. The updated weight $w^t$ is a combination of the current fitting error and the historical weights $w^{t-1}$,
\begin{equation}
    w_{ij}^t=\begin{cases}
        w_{ij}^{t-1}, & d_{chord}(\mathbf{\tilde{G}}_{ij},\mathbf{\hat{G}}_{ij})<\tau\\
        \frac{\tau}{d_{chord}(\mathbf{\tilde{G}}_{ij},\mathbf{\hat{G}}_{ij})}w_{ij}^{t-1}, & d_{chord}(\mathbf{\tilde{G}}_{ij},\mathbf{\hat{G}}_{ij})>\tau
    \end{cases},
\end{equation}
where $\tau$ is the median chordal distance of all available pairs.

The reweighting operation is performed after every $\lambda$ iteration, excluding the first $\lambda_{warm}$ iterations. This initial warm-up period allows the network to converge to a stable state, as reweighting a chaotic network would be futile. All entries' weights are initialized to 1 at the beginning.

\subsection{Loss Function}
\label{subsec:loss}
To obtain the estimated matrix $\mathbf{\hat{G}}$, we penalize the weighted error between the observed matrix $\mathbf{\tilde{G}}$ and the corresponding portion in our generated matrix,
\begin{equation}
    \mathcal{L}=w|\mathbf{\hat{G}}\odot\Omega-\mathbf{\tilde{G}}\odot\Omega|,
\end{equation}
where the $|\cdot|$ is entry-wise $\ell_1$ loss. Consistent with prior work~\cite{tejus2023rotation}, we employ $\ell_1$ loss instead of $\ell_2$ loss. This choice is driven by the robustness of $\ell_1$ loss to outliers, as observed matrices in the multiple rotation averaging problem often contain more contamination compared to matrices in image processing or recommendation domains.

Furthermore, we experimented with the $\ell_{\frac{1}{2}}$ loss, which has been suggested as more suitable for the multiple rotation averaging problem in~\cite{chatterjee2017robust}. However, our experiments revealed that the $\ell_{\frac{1}{2}}$ loss function often leads to instability in the optimization process. As a result, we opted to implement a reweighting scheme to achieve a similar outcome.

\newcommand{\oom}{OOM$^{\dag}$}

\begin{table}[t]
\caption{Results on 1DSfM, ETH, and Stanford 3D. mean angular error/median angular error. Bold and underline denote best and second best performance. $^\dag$Out of memory.}
\label{tab:1dsfm}
\centering
\resizebox{\linewidth}{!}{
\begin{tabular}{cccccccc}
\hline
Dataset & IRLS-$\ell_{\frac{1}{2}}$ & \begin{tabular}[c]{@{}c@{}}EIG\\ IRLS\end{tabular} & CEMP & MPLS & HARA & \begin{tabular}[c]{@{}c@{}}DMF\\ SYNCH\end{tabular} & Ours \\
\hline
ALM  & 4.6/1.4 & \underline{3.8}/1.2 & 4.0/1.7 & 3.9/1.5 & \textbf{3.5}/\underline{1.1} & 4.1/1.2 & \textbf{3.5}/\textbf{1.0} \\
ELS  & 3.3/1.2 & 3.0/0.7 & 3.5/1.1 & 3.5/1.0 & \textbf{2.1}/\underline{0.5} & 2.8/0.8 & \underline{2.2}/\textbf{0.4} \\
GDM & 49.2/28.3& 61.4/70.2& 43.7/8.3 & \underline{41.2}/8.0 & 44.2/\textbf{3.1} & 46.2/10.5& \textbf{39.0}/\underline{6.2} \\
MDR & 7.8/3.2 & 9.3/4.0 & 9.2/5.3 & 6.2/2.0 & \textbf{4.8}/\underline{1.3} & 8.2/2.3 & \underline{6.0}/\textbf{0.9} \\
MND & 1.6/0.7 & 1.8/0.6 & 1.6/0.8 & 1.4/0.6 & \textbf{1.1}/\underline{0.5} & 1.5/0.6 & \underline{1.4}/\textbf{0.4} \\
ND1 & 3.8/1.0 & 3.6/\underline{0.7} & 3.0/1.0 & 2.7/\underline{0.7} & \textbf{1.6}/\textbf{0.6} & 3.8/0.8 & \underline{1.8}/\textbf{0.6} \\
NYC & \underline{3.2}/1.4 & 3.5/1.8 & 3.4/1.6 & \underline{3.2}/1.3 & \textbf{3.1}/\underline{1.3} & 3.6/1.8 & \underline{3.2}/\textbf{1.1} \\
PDP & 4.9/2.6 & \underline{3.9}/1.0 & 4.7/2.3 & \underline{3.9}/1.9 & \textbf{3.3}/\underline{0.9} & 4.8/1.0 & \underline{3.9}/\textbf{0.8} \\
PIC & 5.9/2.7 & 8.0/4.7 & \oom & \oom            & \textbf{4.4}/\underline{2.2} & 7.1/4.5 & \underline{4.5}/\textbf{1.9} \\
ROF & 3.1/1.7 & 24.6/2.7 & 3.5/1.7 & 2.8/\underline{1.4} & \underline{2.7}/1.5 & 3.1/1.8 & \textbf{2.6}/\textbf{1.3} \\
TOL & 4.0/2.4 & 4.4/2.8 & \underline{3.7}/\textbf{1.9} & 3.9/2.4 & 4.2/2.7 & 3.8/2.7 & \textbf{3.6}/\underline{2.1} \\
TFG & \underline{3.6}/2.1 & 73.8/21.7& \oom   & \oom      & \textbf{3.5}/\underline{1.9} & \oom      & \textbf{3.5}/\textbf{1.6} \\
USQ & 7.3/3.9 & 6.6/4.7 & 6.2/\underline{3.3} & 6.2/3.6 & \underline{5.7}/3.9 & 6.2/4.4 & \textbf{5.5}/\textbf{3.2} \\
VNC & 10.9/4.5 & 8.6/1.6 & 7.3/3.0 & 7.3/2.8 & \textbf{6.2}/\underline{1.4} & 8.4/1.6 & \underline{6.3}/\textbf{1.2} \\
YKM & 3.8/1.7 & 3.8/1.8 & 3.6/\underline{1.5} & 3.6/1.6 & \underline{3.0}/1.6 & 3.6/1.7 & \textbf{2.9}/\textbf{1.4} \\
\hline
Gaz. S & 2.5/1.7 & \textbf{2.3}/\underline{1.5} & \textbf{2.3}/1.6 & 2.5/1.7 & \underline{2.4}/\textbf{1.4} & 12.1/4.1 & 2.5/\underline{1.5} \\
Gaz. W & 2.3/\underline{1.6} & \textbf{2.1}/\textbf{1.4} & 2.6/1.8 & 2.3/1.8 & \underline{2.2}/\underline{1.6} & 3.6/\underline{1.6} & \textbf{2.1}/1.7 \\
Wood A & 7.5/6.2 & 47.5/2.2 & 6.5/5.3 & 9.7/9.8 & \textbf{4.5}/\textbf{1.8} & 20.5/9.8 & \underline{5.5}/\underline{2.0} \\
Wood S & 8.0/5.8 & \underline{6.6}/\underline{3.7} & 11.5/9.7 & 9.7/7.8 & 57.7/10.3 & 25.1/13.1& \textbf{6.1}/\textbf{3.4} \\
\hline
Arm. & 8.0/8.8 & 8.3/8.1 & 3.3/\underline{2.4} & 5.2/4.3 & \underline{3.0}/3.7 & 11.1/10.1& \textbf{0.4}/\textbf{0.4} \\
Bud.    & \textbf{4.3}/\textbf{4.3} & 7.0/5.6 & 7.4/5.8 & 6.8/6.3 & \underline{6.0}/\underline{4.5} & 7.3/6.3 & 12.4/8.9 \\
Bun.     & 3.1/2.0 & 5.4/3.6 & 2.1/1.5 & 2.5/1.6& \textbf{0.3}/\textbf{0.2} & 11.4/10.1& \underline{0.6}/\underline{0.3} \\
Dra.    & \textbf{9.3}/\underline{9.2} & 9.7/10.1& 10.7/\textbf{9.1} & \underline{9.5}/9.7 & 9.6/9.4 & \textbf{9.3}/10.2& 10.0/10.7 \\
\hline
\end{tabular}
}
\end{table}

\section{Experiments}
\label{sec:experiments}

\subsection{Baseline and Metric}

We compare our method with the following methods: IRLS-$\ell_{\frac{1}{2}}$~\cite{chatterjee2017robust}, EIG-IRLS~\cite{arrigoni2016spectral}, CEMP~\cite{lerman2022robust}, MPLS~\cite{shi2020message}, HARA~\cite{lee2022hara}, DMF-SYNCH~\cite{tejus2023rotation}. The code of these algorithms are released by the authors.

To evaluate the accuracy, the predicted absolute camera orientations $\mathbf{R}^{pred}$ are compared with ground truth camera orientations $\mathbf{R}^{gt}$ in terms of mean(mn) and median(md) of angular errors $\mathcal{A}=\{d(\mathbf{R}^{pred}_i, \mathbf{R}^{gt}_i\mathbf{R}_{align})\}_{i=1}^{N}$. It is worth noting that the output camera rotations need to align with the ground truth by $\mathbf{R}_{align}$ to resolve the gauge ambiguity. 

\subsection{Results}
All quantitative experiment results are shown in Tab. \ref{tab:1dsfm}.

1DSfM dataset~\cite{wilson2014robust} contains 15 outdoor scenes with ground-truth camera poses and relative orientations computed by Bundler~\cite{snavely2006photo}, in our experiments, only the cameras with ground-truth orientations and the edges between these cameras are used. It can be seen that our approach achieves great performance in almost all scenes, with particular strength in the median metric where we achieved the best results on 13 of the 15 scenarios.

ETH dataset~\cite{pomerleau2012challenging} is an outdoor point cloud dataset containing 4 scenes, each scene has about 33 scans, and the ground truth orientations of scans are provided. The input pose graphs are based on the pipeline in SGHR~\cite{wang2023robust}. Our method continues to provide impressive results on this dataset, particularly on the challenging scene Wood summer.

Stanford 3D dataset ~\cite{curless1996volumetric} is a famous object-level point cloud dataset, each object contains 10 to 15 frames and the ground truth orientations are also provided.

\begin{figure}[ht]
    \centering
    \begin{subfigure}{0.23\linewidth}
        \centering
        \includegraphics[width=\linewidth]{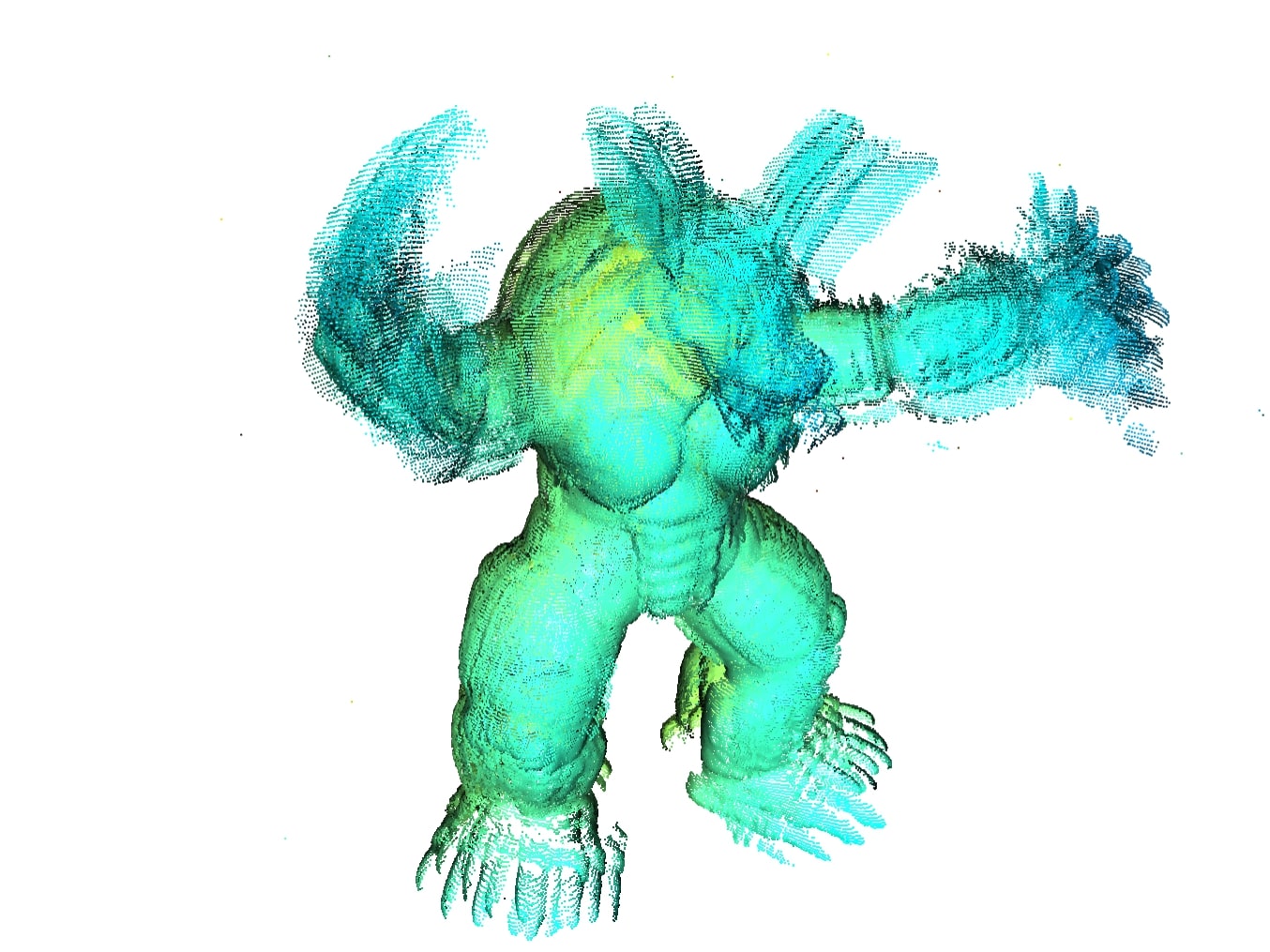}
        \caption{IRLS-$\ell_{\frac{1}{2}}$}
    \end{subfigure}
    \begin{subfigure}{0.23\linewidth}
        \centering
        \includegraphics[width=\linewidth]{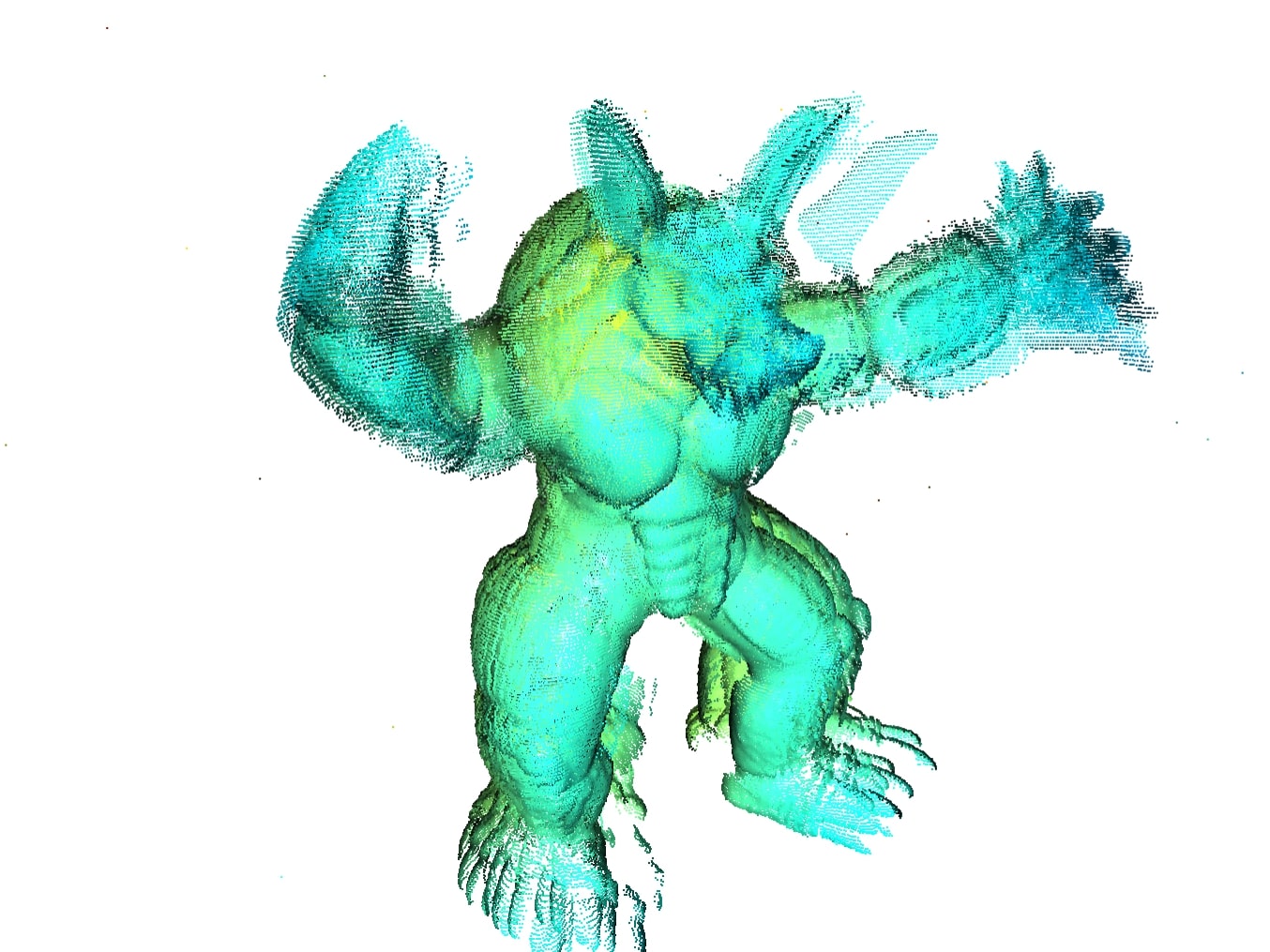}
        \caption{EIG-IRLS}
    \end{subfigure}
    \begin{subfigure}{0.23\linewidth}
        \centering
        \includegraphics[width=\linewidth]{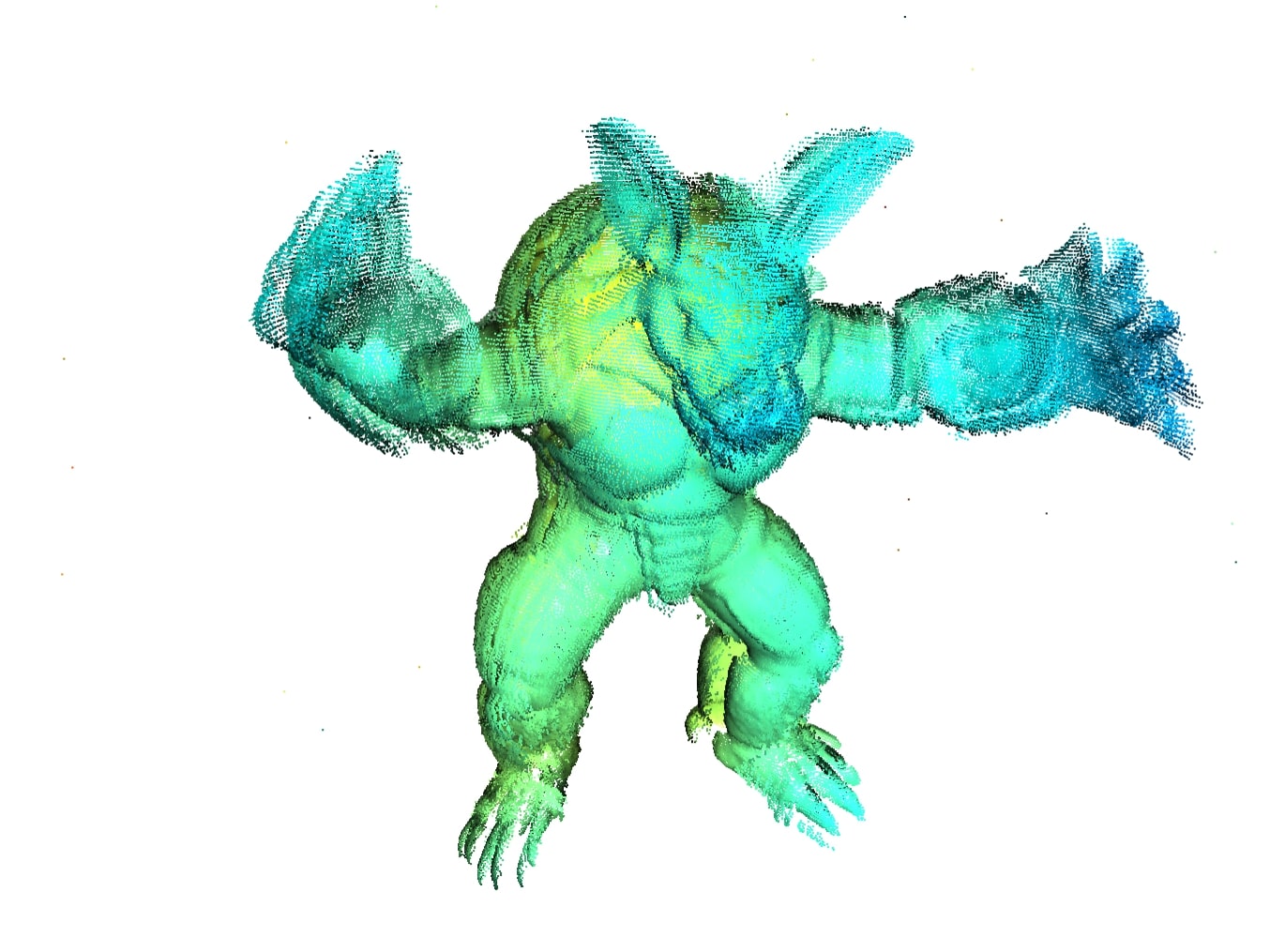}
        \caption{CEMP}
    \end{subfigure}
    \begin{subfigure}{0.23\linewidth}
        \centering
        \includegraphics[width=\linewidth]{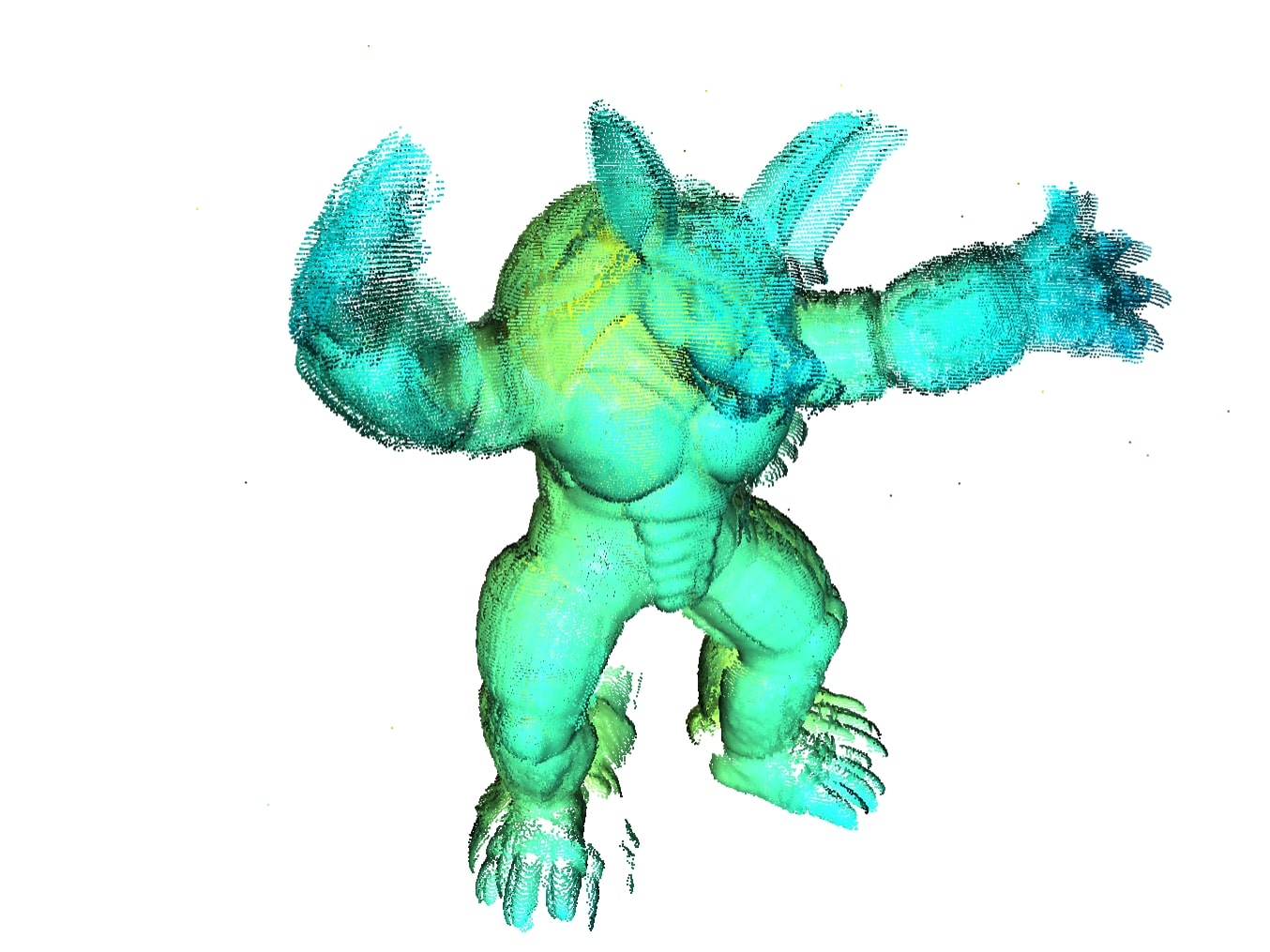}
        \caption{MPLS}
    \end{subfigure}

    \begin{subfigure}{0.23\linewidth}
        \centering
        \includegraphics[width=\linewidth]{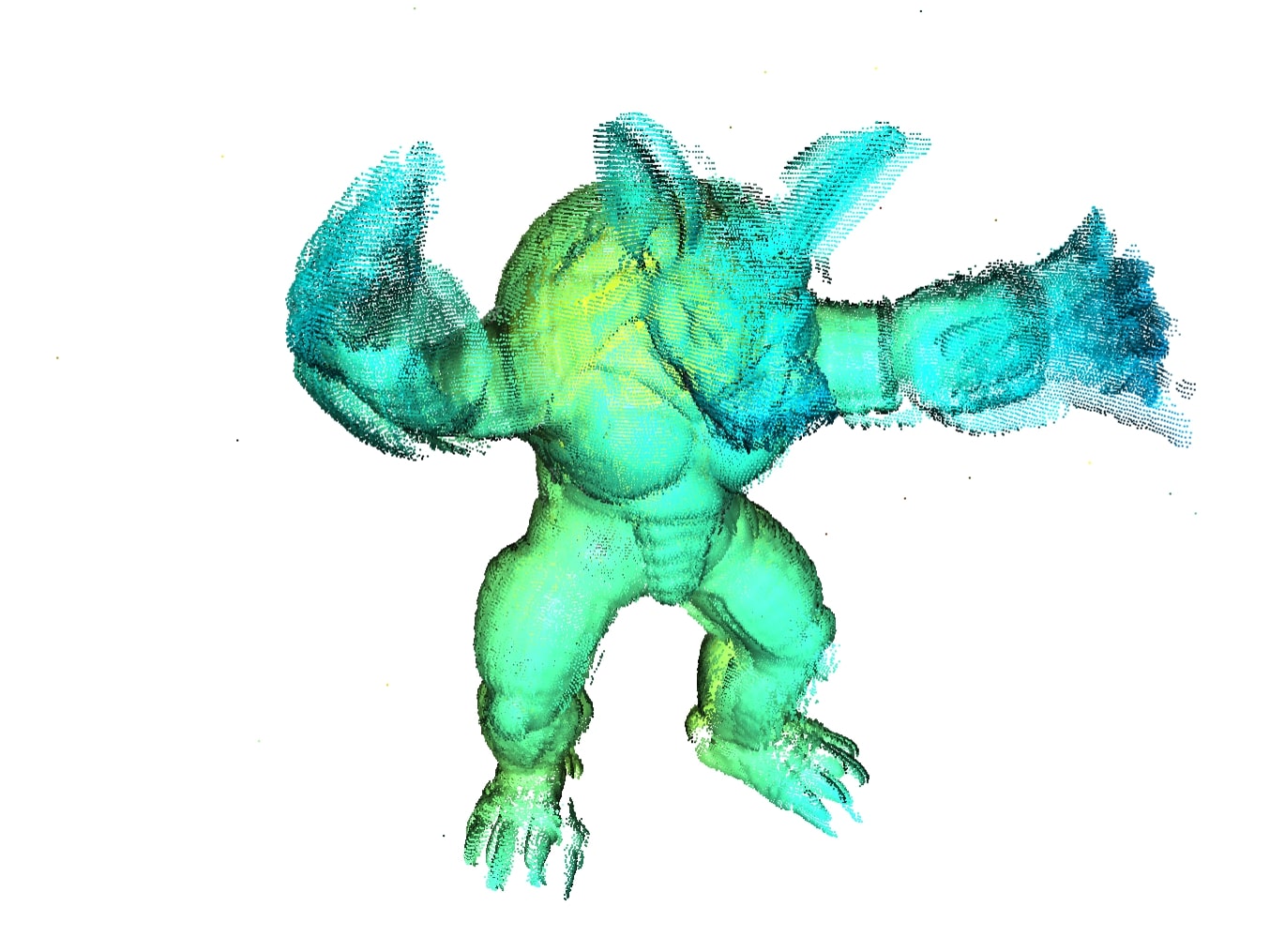}
        \caption{HARA}
    \end{subfigure}
    \begin{subfigure}{0.23\linewidth}
        \centering
        \includegraphics[width=\linewidth]{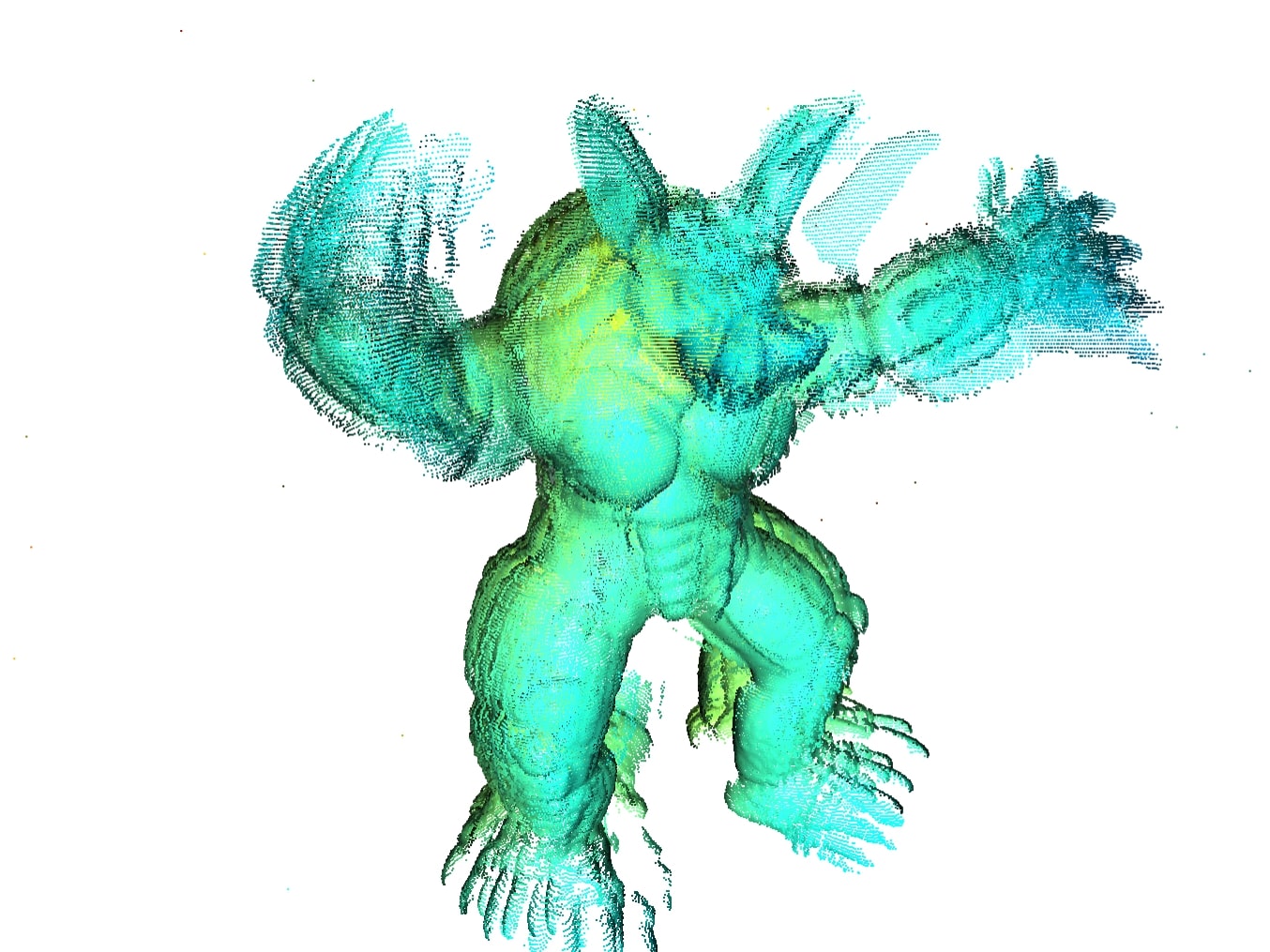}
        \caption{DMF}
    \end{subfigure}
    \begin{subfigure}{0.23\linewidth}
        \centering
        \includegraphics[width=\linewidth]{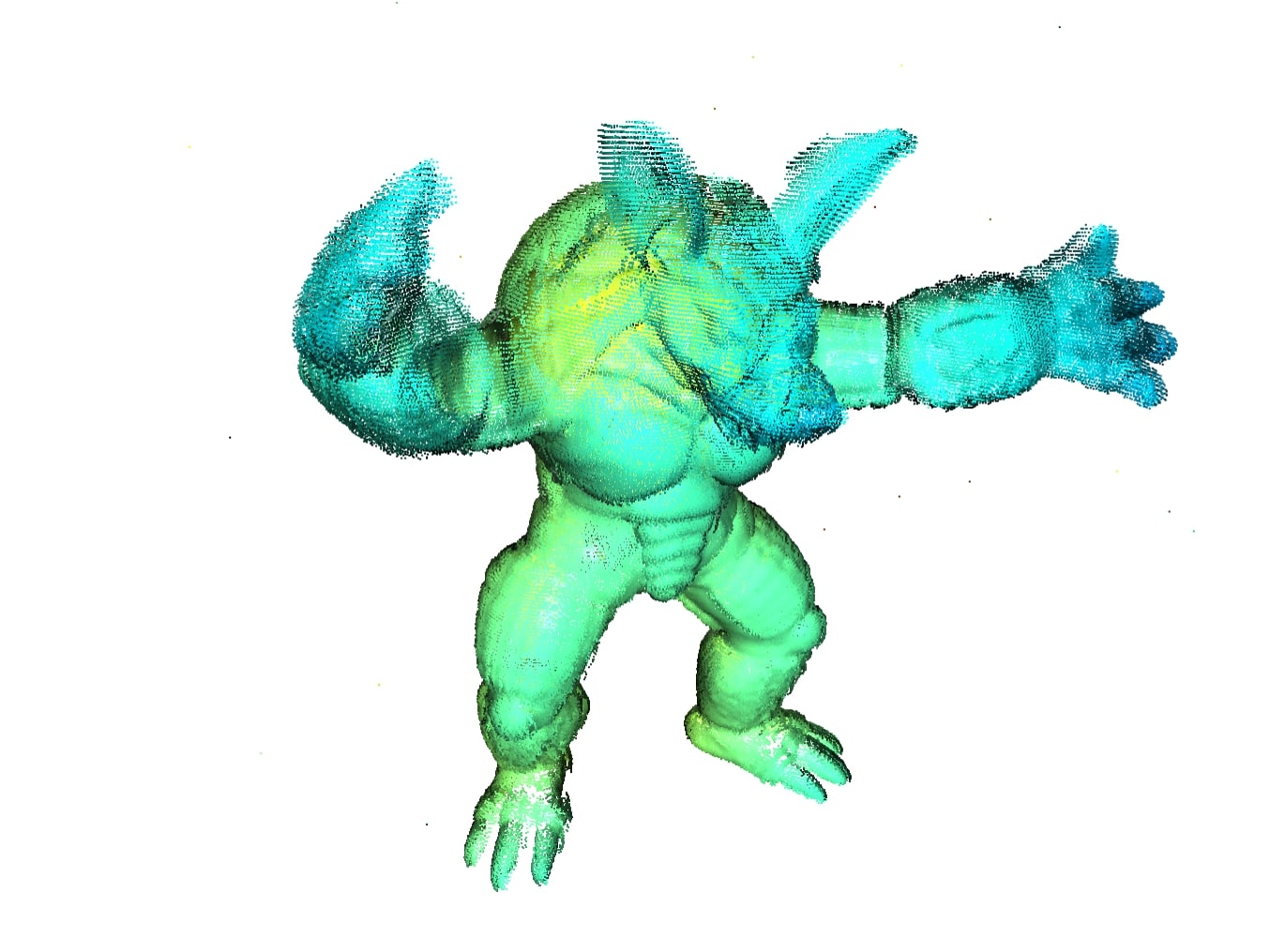}
        \caption{Ours}
    \end{subfigure}
    \begin{subfigure}{0.23\linewidth}
        \centering
        \includegraphics[width=\linewidth]{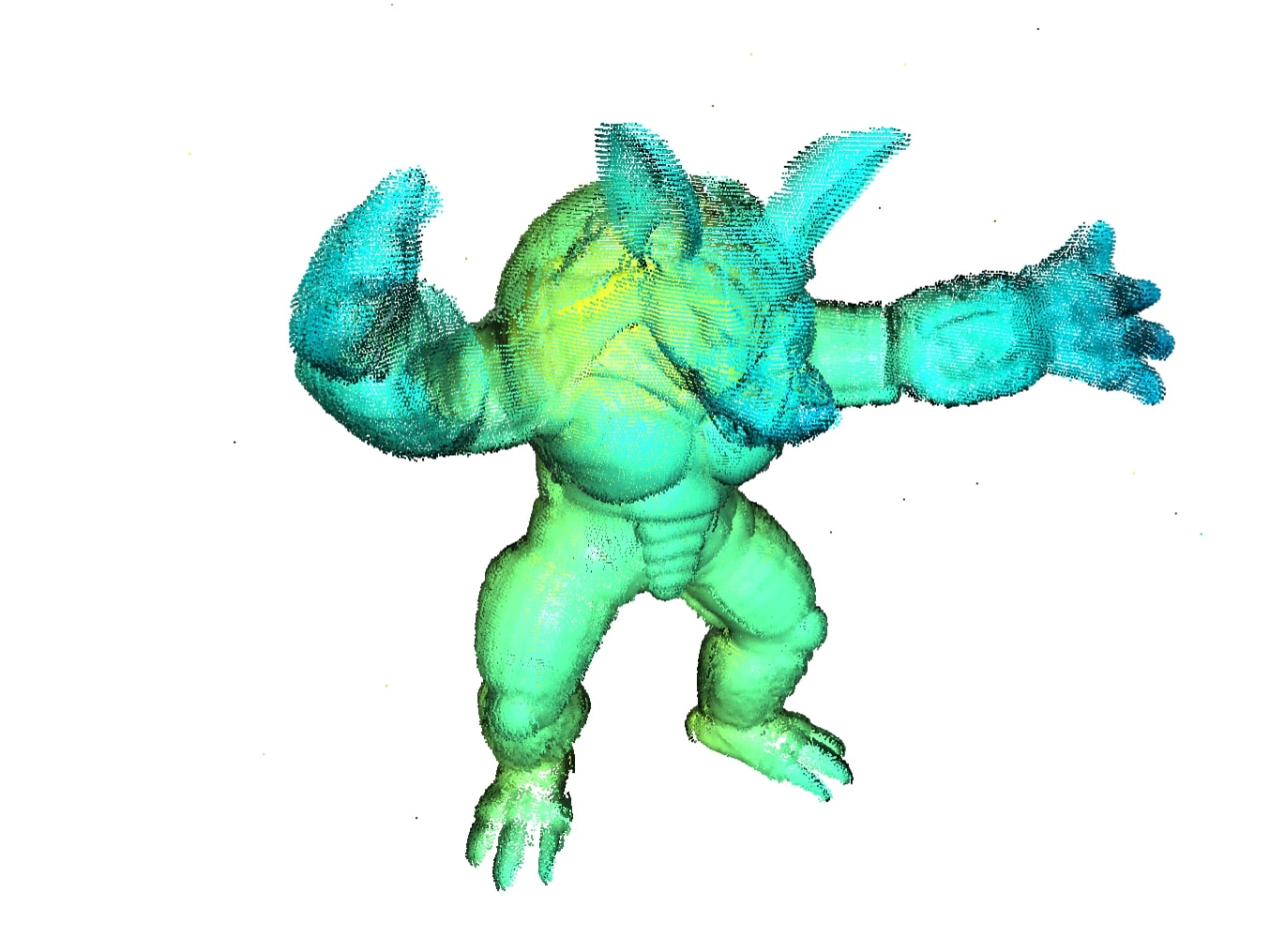}
        \caption{GT}
    \end{subfigure}
    \caption{Visualization on Stanford 3D Armadillo.}
    \label{fig:arm}
\end{figure}

Following the protocol in prior works~\cite{govindu2013averaging, arrigoni2016global}, we use Generalized-ICP~\cite{segal2009generalized} to obtain pairwise registration results.
We can see that our method also provides distinguished or comparable results in this experiment, especially on Armadillo and Bunny. Qualitative results are shown in Fig. \ref{fig:arm}.

\subsection{Ablation Study}
To assess the effectiveness of each component, we conduct ablation studies on the 1DSfM dataset. Tab. \ref{tab:ablation} provides a summary of the ablation results. We first assess the impact of spanning tree-based edge filtering in Exp. (a). Additionally, we evaluate a spanning tree based solely on the mean triplet error in Exp. (b), highlighting the importance of support count $s$, particularly on the mean error metric. In Exp. (c), we remove the low-rank and symmetric constraints and use a vanilla network architecture, the results illustrate that our bottleneck design effectively prevents network overfitting. Finally, in Exp. (d), we demonstrate that the reweighting scheme can further enhance accuracy.

\begin{table}[ht]
\centering
\caption{Ablation study on 1DSfM dataset.}
\resizebox{\linewidth}{!}{
\begin{tabular}{cccccc}
\hline
Dataset& \begin{tabular}[c]{@{}c@{}}(a) w/o\\ edge \\ filter\end{tabular} & \begin{tabular}[c]{@{}c@{}}(b) w/o\\ support \\ count\end{tabular} & \begin{tabular}[c]{@{}c@{}}(c) w/o\\ explicit \\constraint\end{tabular} & \begin{tabular}[c]{@{}c@{}}(d) w/o\\ reweighing \\ scheme\end{tabular} & \begin{tabular}[c]{@{}c@{}}full \\ modules\end{tabular} \\
\hline
ALM     & 3.9/\textbf{1.0} & \textbf{3.4}/\textbf{1.0} & 5.6/1.1 & 3.7/1.2 & 3.5/\textbf{1.0} \\
ELS     & 2.3/0.5 & \textbf{2.2}/0.5 & 3.2/1.1 & 2.5/0.7 & \textbf{2.2}/\textbf{0.4} \\
GDM     & \textbf{37.5}/6.6 & 38.4/\textbf{2.8} & 42.5/12.3& 38.5/6.9 & 39.0/6.2 \\
MDR     & 7.2/1.1 & 8.5/\textbf{0.9} & 10.8/5.8 & 6.9/1.8 & \textbf{6.0}/\textbf{0.9} \\
MND     & \textbf{1.2}/0.5 & 2.6/0.5 & 2.8/0.6 & 1.5/0.6 & 1.4/\textbf{0.4} \\
ND1     & 3.6/\textbf{0.6} & 3.1/\textbf{0.6} & 2.4/\textbf{0.6} & 2.0/0.7 & \textbf{1.8}/\textbf{0.6} \\
NYC     & 3.3/1.1 & \textbf{2.6}/\textbf{1.0} & 4.3/2.0 & 3.4/1.5 & 3.2/1.2 \\
PDP     & 5.1/\textbf{0.8} & 4.2/\textbf{0.8} & 4.8/1.1 & 4.1/0.9 & \textbf{3.9}/\textbf{0.8} \\
PIC     & 4.9/2.2 & 8.1/\textbf{1.8} & 22.5/2.5 & 5.2/2.7 & \textbf{4.5}/1.9 \\
ROF     & 2.7/\textbf{1.3} & 53.1/1.9 & 7.4/2.1 & 2.9/1.5 & \textbf{2.6}/\textbf{1.3} \\
TOL     & 3.8/\textbf{2.1} & 3.9/\textbf{2.1} & 4.7/2.7 & 3.9/2.7 & \textbf{3.6}/\textbf{2.1} \\
TFG     & 3.6/1.8 & 21.2/16.3 &  \oom  & 3.9/2.1 & \textbf{3.5}/\textbf{1.6} \\
USQ     & 5.8/3.3 & 5.7/3.2 & 8.7/\textbf{2.8} & 6.0/4.1 & \textbf{5.5}/3.2 \\
VNC     & 8.3/\textbf{1.2} & 14.0/1.3 & 14.7/1.3 & 6.5/1.5 & \textbf{6.3}/\textbf{1.2} \\
YKM     & 3.2/\textbf{1.4} & 3.5/\textbf{1.4} & 4.5/1.8 & 3.1/1.7 & \textbf{2.9}/\textbf{1.4} \\

\hline
\end{tabular}
}
\label{tab:ablation}
\end{table}

\section{Conclusion}
\label{sec:conclusion}
In this paper, we introduce a novel method for solving the multiple rotation averaging problem. Our approach begins with the utilization of a spanning tree-based edge filtering to mitigate the influence of outliers in input relative rotations. We then employ an explicitly low-rank and symmetric linear neural network to directly recover absolute orientations in a deep matrix factorization manner. Our model achieves unsupervised learning by relying solely on observed relative rotations as guidance. Additionally, we apply a reweighting scheme and a dynamic depth selection mechanism to further enhance the robustness. Experiments on various datasets validate the effectiveness of our method.

\newpage







\bibliography{main}

\end{document}